**Uncertainty quantification for improving radiomic-based models in radiation pneumonitis prediction.**

**Chanon Puttanawarut[1,2] (Corresponding author)**

[1] Chakri Naruebodindra Medical Institute, Faculty of Medicine, Ramathibodi Hospital, Mahidol University, 111 Suvarnabhumi Canal Road, Bang Pla Subdistrict, Bang Phli, Samut Prakan, Thailand

[2] Department of Clinical Epidemiology and Biostatistics, Faculty of Medicine, Ramathibodi Hospital, Mahidol University, 270 Rama VI Road, Thung Phayathai Subdistrict, Ratchathewi District, Bangkok, Thailand

E-mail: chanonp@protonmail.com

**Romen Samuel Wabina[2]**

[2] Department of Clinical Epidemiology and Biostatistics, Faculty of Medicine, Ramathibodi Hospital, Mahidol University, 270 Rama VI Road, Thung Phayathai Subdistrict, Ratchathewi District, Bangkok, Thailand

E-mail: romensamuel.wab@mahidol.edu

**Nat Sirirutbunkajorn[3] (Corresponding author)**

[3] Department of Diagnostic and Therapeutic Radiology, Faculty of Medicine, Ramathibodi Hospital, Mahidol University, 270 Rama VI Road, Thung Phayathai Subdistrict, Ratchathewi District, Bangkok, Thailand

E-mail: nat19012537@gmail.com

## Abstract

**Background and purpose**: Radiation pneumonitis is a side effect of thoracic radiation therapy. Recently, machine learning models with radiomic features have improved radiation pneumonitis prediction by capturing spatial information. To further support clinical decision-making, this study explores the role of post hoc uncertainty quantification methods in enhancing model uncertainty estimate.
**Materials and methods**: We retrospectively analyzed a cohort of 101 esophageal cancer patients. This study evaluated four machine learning models: logistic regression, support vector machines, extreme gradient boosting, and random forest, using 15 dosimetric, 79 dosiomic, and 237 radiomic features to predict radiation pneumonitis. We applied uncertainty quantification methods, including Platt scaling, isotonic regression, Venn-ABERS predictor, and conformal prediction, to quantify uncertainty. Model performance was assessed through an area under the receiver operating characteristic curve (AUROC), area under the precision-recall curve (AUPRC), and adaptive calibration error using leave-one-out cross-validation.
**Results**: Highest AUROC is achieved by the logistic regression model with the conformal prediction method (AUROC 0.75±0.01, AUPRC 0.74±0.01) at a certainty cut point of 0.8. Highest AUPRC of 0.82±0.02 (with AUROC of 0.67±0.04) achieved by The extreme gradient boosting model with conformal prediction at the 0.9 certainty threshold. Radiomic and dosiomic features improve both discriminative and calibration performance.
**Conclusions**: Integrating uncertainty quantification into machine learning models with radiomic and dosiomic features may improve both predictive accuracy and calibration, supporting more reliable clinical decision-making. The findings emphasize the value of uncertainty quantification methods in enhancing applicability of predictive models for radiation pneumonitis in healthcare settings.

## 1. Introduction

Radiation pneumonitis (RP) is a common side effect of thoracic radiation therapy, characterized by inflammation of the lungs resulting from radiation exposure. The timing and severity of RP can vary widely among patients, but it is typically detected within the first 8 months after radiation [1]. The incidence rate ranges from 15–40% among patients receiving thoracic radiation [2]. Recently, machine learning (ML) has been used to predict RP using features from dose-volume histograms (DVHs) and clinical data [2]. However, DVH-based features lack spatial information. To address this, radiomics and dosiomics, spatially informed features, have been developed and shown to improve prediction performance [3–6].

From previous studies, ML models with radiomics and/or dosiomics are commonly evaluated based on their discriminative ability. However, high discriminative performance alone is insufficient for clinical applications since it does not guarantee robustness, generalizability across diverse patient populations, or practical integration into clinical workflows. In the medical field, inaccurate or overly confident predictions can lead to harmful consequences, such as misdiagnosis, inappropriate treatment decisions, or delayed interventions that compromise patient safety and outcomes. For a general classification model, we can view probability output as an uncertainty estimate. However, a model might have good discriminative ability but exhibit inaccurate confidence levels [7–9]. This is where uncertainty quantification (UQ) plays a crucial role. UQ helps assess not only whether a model is correct but also how certain it is in its predictions, thereby promoting reliable and trustworthy application of AI in healthcare settings [10–13]. Furthermore, previous studies show that incorporating UQ methods can help improve model discriminative performance and clinical decision support in various ML medical tasks such as Alzheimer's disease prediction [14], diabetic retinopathy detection [15], and polyp classification [16].

A recent review of UQ in radiotherapy identified its applications in image synthesis, registration, contouring, dose prediction, and outcome prediction [17]. For outcome prediction, UQ has been applied to tasks like local control prediction [18], survival prediction [19] and locoregional recurrence prediction [20]. For RP prediction, existing studies [18,21] have focused on improving uncertainty evaluation metrics but have not explicitly demonstrated how uncertainty enhances discriminative performance.

In this study, we aim to explore the impact of integrating UQ into widely used ML models that leverage radiomic and dosiomic features for RP prediction in esophageal cancer patients. While prior studies have employed Bayesian networks [21] and Gaussian processes integrated with deep neural networks [18] for UQ in RP prediction, these methods require integration into the model during training. This requirement renders them unsuitable for existing ML models, highlighting the need for alternative methods that can seamlessly retrofit UQ capabilities into pre-existing frameworks. Our focus will be on post hoc (adjustments occur after initial model training) UQ methods since it can easily be integrated into the existing common radiomic based ML model in RP without requiring extensive modifications. Additionally, we evaluate the models using both discriminative and uncertainty evaluation metrics, and we assess how incorporating uncertainty can enhance discriminative performance.

## 2. Material and Methods

### 2.1 **Dataset**

This study was approved by the Institutional Review Board (IRB) of Ramathibodi hospital, Bangkok, Thailand approval (MURA2024/933) in accordance with the Declaration of Helsinki. This study included esophageal cancer patients aged over 15 years who underwent thoracic radiation therapy for esophageal cancer, regardless of the indication (pre-operative concurrent chemoradiation, definitive chemoradiation, post-op radiation or palliative radiation) between January 2011 and June 2019.

A radiation oncologist reviewed and extracted clinical data for each patient from electronic medical records. For this study, the positive class was defined as grade 1 or more radiation pneumonitis. After exclusion criteria, 101 patients were eligible and included in the final analysis. For more details of dataset grading of RP and exclusion criteria, I refer to **Supplementary A**. Patient and treatment characteristics are summarized in **Supplementary Table 1**, **2**.

### 2.2 **Preprocessing and Features**

Prior to radiotherapy, pretreatment CT images were acquired using the Optima 580 CT simulator (GE Healthcare, Milwaukee, WI, USA) prior to radiotherapy treatment. The total dose distribution was converted to an equivalent dose of 2 Gy (EQD2). From now on the dose distribution will refer to dose distribution in EQD2. The dose distributions and pretreatment CT images were then resampled to have voxel size of $1.5 \times 1.5 \times 1.5$ mm$^3$ using b-spline algorithm. A Gaussian filter was applied to the CT image before resampling as an anti-aliasing filter [23,24]. The regions of interests (ROIs) were resampled to match the pretreatment CT images using the nearest neighbor algorithm.

In this study, we extracted three categories of features: dosimetric, dosiomic (derived from the dose distribution), and radiomic (derived from the pretreatment CT images). Dosimetric features included mean lung dose and relative lung volume receiving more than a specific dose threshold x (Vx), for x in [5, 10, …, 70]. Dosiomic and radiomic features were extracted using the PyRadiomics v3.0.1 [25]. PyRadiomics adheres to the Imaging Biomarker Standardization Initiative (IBSI) guidelines [26] in most aspects, although certain deviations exist, as detailed in the PyRadiomics documentation [27].

The dosiomic feature set included 18 first-order statistics and 61 texture features, computed from dose distribution. Radiomic features were extracted from pretreatment CT images within three dose-based lung ROIs, defined by thresholds of 10 Gy, 15 Gy, and 20 Gy, similar to previous study [28]. The same set of first-order and texture features used for dosiomics was extracted for each ROI. Unless otherwise specified, all other feature extraction parameters followed the default settings of PyRadiomics v3.0.1.

In total, each patient comprised 15 dosimetric, 79 dosiomic, and 237 radiomic features. All features (dosimetric, dosiomic, and radiomic) then were standardized to a scale of 0 to 1 by min-max normalization. To reduce redundancy, Spearman's rank correlation test was employed to identify the correlation between all possible pairs of features. If a correlation exceeding 0.8 was identified, we removed the feature that had the highest number of correlations exceeding 0.8 with other features. Both normalization parameters and feature selection were determined using the training set within a nested cross-validation framework (**Figure 1**). For more details about preprocessing and features extraction we refer to **Supplementary B**.

### 2.3 Training process

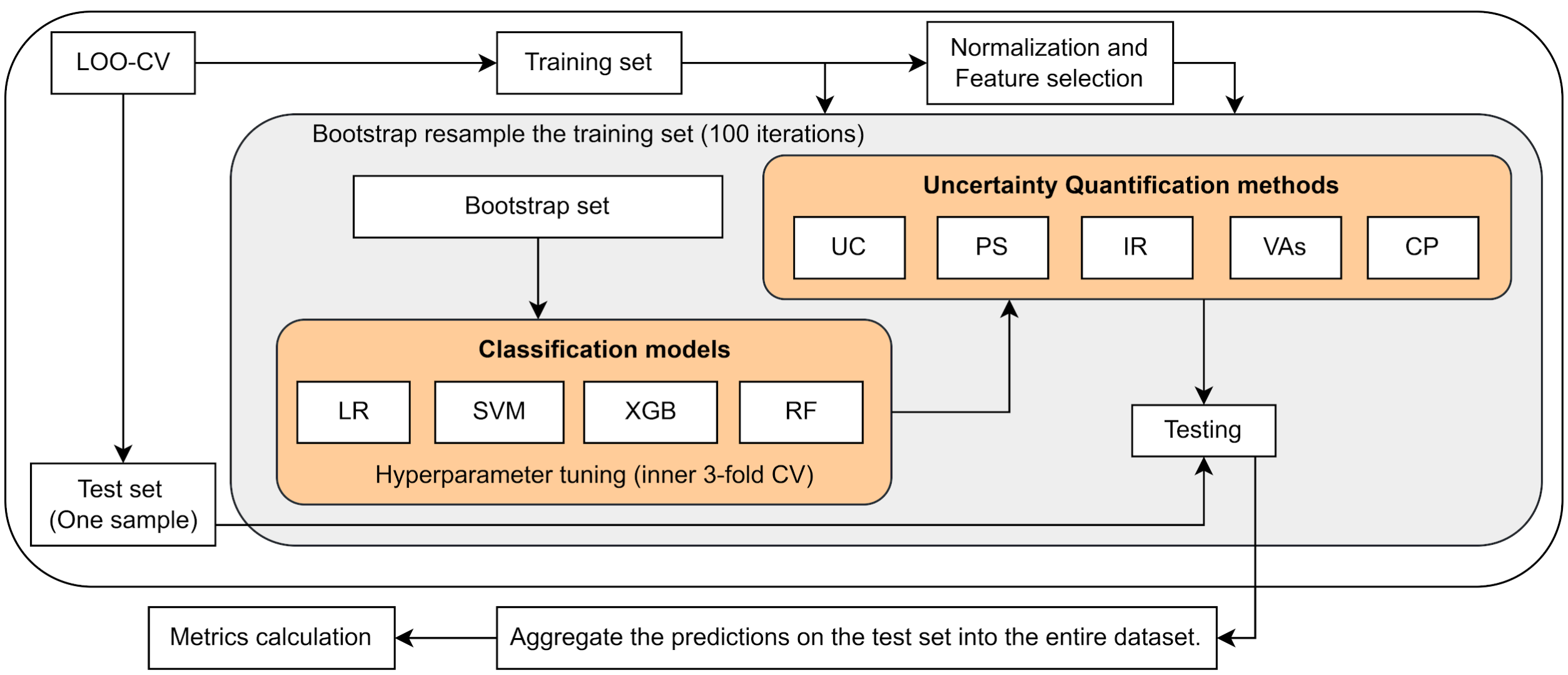

**Figure 1**: This figure depicts the study's workflow for classification and uncertainty quantification. Initially, LOO-CV splits the dataset into a single test sample and a training set. Normalization and feature selection are performed on the training set, which is then bootstrapped (100 iterations) to bootstrap sets. Four classification models, LR, SVM, XGB, and RF, are trained and hyperparameter-tuned on these bootstrap sets. Five uncertainty quantification methods (UC, PS, IR, VAs, CP) are applied. The test sample is used to generate predictions for each combination of model, uncertainty method, and bootstrap iteration. These predictions are then aggregated into the entire dataset for each combination, and evaluation metrics are calculated. Abbreviations: LR: logistic regression; SVM: support vector machine; XGB: extreme gradient boosting; RF: random forest; UC: uncalibrated; PS: Platt scaling; IR: isotonic regression; VAs: Venn-ABERS; CP: conformal prediction; LOO-CV: leave-one-out cross-validation

Figure 1 outlines the training pipeline. Data is split using leave-one-out cross-validation (LOO-CV), where each sample is tested once. To improve robustness, 100 bootstrap samples are generated from the training data. Four models are used for classification: logistic regression (LR), support vector machines (SVM), random forest (RF), and extreme gradient boosting (XGB). LR uses L1 or L2 regularization, selected via hyperparameter tuning. The selection of these models was informed by a recent systematic review of radiomic- and dosiomic-based ML models for RP prediction [3] and broader reviews of radiomics literature [29], which showed LR, SVM, and RF among the most widely used approaches, with XGB added as a modern alternative.

Hyperparameters are tuned using nested three-fold cross-validation within each bootstrap set. UQ methods, uncalibrated (UC), Platt scaling (PS) [30], isotonic regression (IR), Venn-ABERS (VAs) [31], and conformal prediction (CP) [32–34], are applied to generate either probability estimates or p-values. Further details of training process, hyperparameters and UQ methods are in **Supplementary C, D, E**. Code is available at: https://github.com/44REAM/RP-Radiomic-Uncertainty.

### 2.4 Evaluations

The evaluation was performed on an entire set that aggregates all test data from each fold of the LOO-CV (**Supplementary D**). For each of the 20 pipelines (combinations of classification model and UQ method), we obtained 100 sets of predictions from the bootstrap iterations. These predictions were then compared to the true labels to assess predictive performance. To assess predictive performance, we calculated the area under the receiver operating characteristic curve (AUROC) and the area under the precision-recall curve (AUPRC). Additionally, we evaluated the UQ methods using uncertainty evaluation metric, specifically a calibration metric, which can be viewed as a form of uncertainty evaluation. Calibration metrics assess the alignment between predicted probabilities and the actual frequency of correct predictions, providing insight into how well the model's predicted confidence reflects reality. We used adaptive calibration error (ACE) [35] as our calibration metric. A lower ACE suggests a better match between the predicted probabilities and actual outcomes, indicating a well-calibrated model while high ACE represents a significant discrepancy between the model’s predicted probabilities and the actual outcomes. We refer to **Supplementary F** for more information about ACE.

To assess the statistical significance of differences in these metrics between methods, we conducted paired t-tests by comparing corresponding bootstrap iterations with different methods. A p-value of less than 0.05 was considered statistically significant.

## 3. **Results**

### 3.1 Uncertainty effect on Prediction performance

To explore the impact of UQ methods, we focused on the performance metrics for the top k% most certain predictions. Specifically, we calculated AUROC and AUPRC (**Figure 2**) progressively, starting from the top 10% of the most certain predictions, incrementally increasing up to 100% of the dataset. This approach enabled us to compute performance metrics at various **coverage levels**, defined as the proportion of the dataset used in the evaluation, ranging from 0.1 to 1. The plots (**Figure 2**) represent median AUROC and AUPRC across 100 bootstrap iterations, with results linearly interpolated at coverage intervals of 0.01 before calculating median for each coverage. By comparing performance across these coverage levels, we gained insights into how UQ methods influence the model's performance. The models analyzed in this section were trained using all feature types (dosimetric, dosiomic, and radiomic). For reference, the corresponding results obtained without bootstrapping are presented in **Supplementary Figure 2**.

For **Figure 2**, in terms of AUROC, the CP methods consistently achieve performance comparable to or better than the UC baseline. The PS method also performs similarly or better than UC across most models, except for the LR model, where its performance is lower. For AUPRC, the IR method demonstrates the best performance overall,  although these improvements come with a decrease in AUROC. It shows consistent improvements across all models, with the exception of the LR model at low coverage levels. It is important to note that in some cases, the coverage curves for IR and VAs do not extend fully to low coverage. This occurs because these methods produce identical predictions for certain samples, resulting in limited coverage ranges.

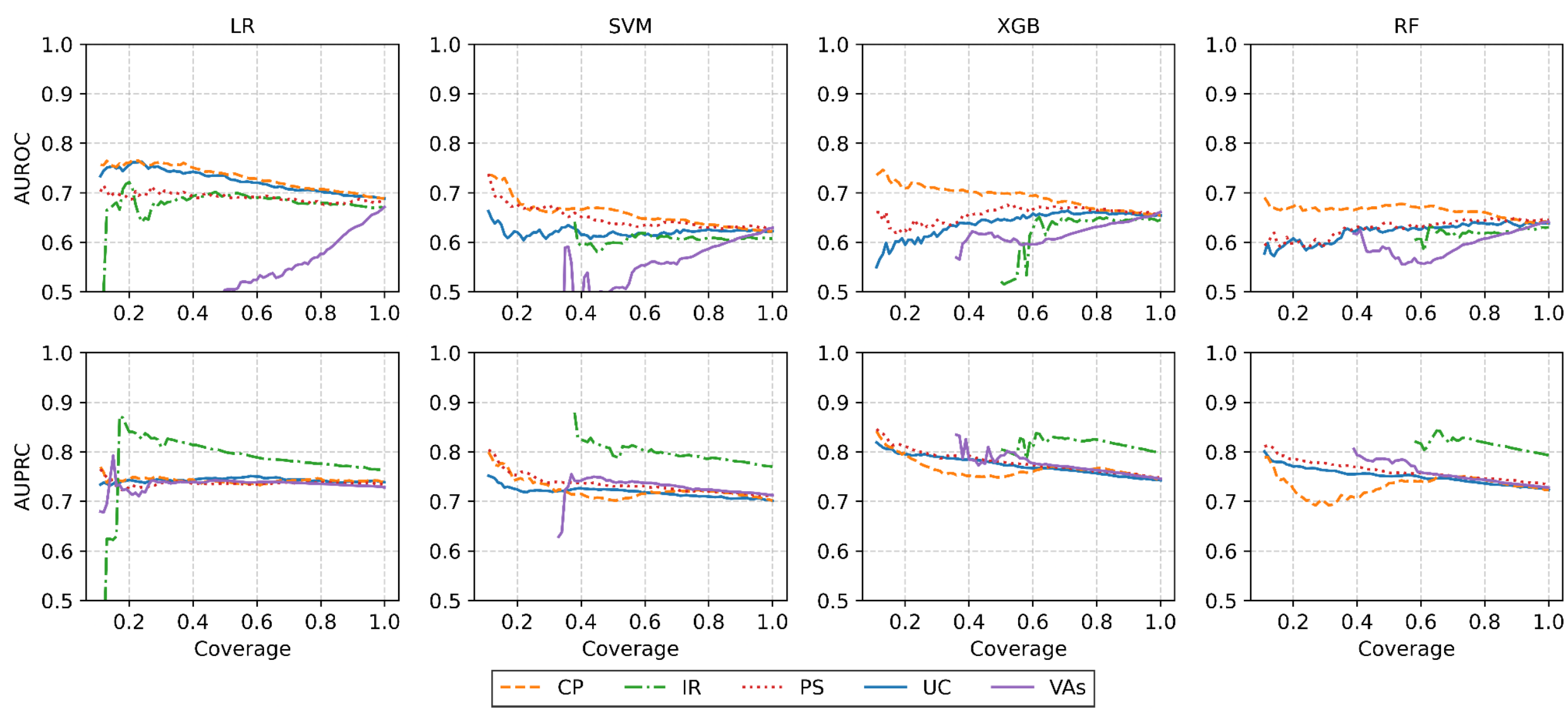

**Figure 2**: Median AUROC and AUPRC across 100 bootstrap results of each ML model with different UQ methods across varying coverage levels. Abbreviations: LR: logistic regression; SVM: support vector machine; XGB: extreme gradient boosting; RF: random forest; UC: uncalibrated; PS: Platt scaling; IR: isotonic regression; VAs: Venn-ABERS; CP: conformal prediction

To better simulate clinical scenarios, where users assess output uncertainty before making decisions, we provided model performance metrics based on specific certainty thresholds in **Table 2**. Users may trust the model's predictions when the certainty is high and may be more skeptical when it is low. Specifically, we evaluated model performance at certainty thresholds of 0.5 (all data), 0.8, and 0.9. Predictions with certainty below these cutoffs were excluded from the evaluation. The corresponding results obtained without bootstrapping are presented in **Supplementary Table 4**.

The results in **Table 2** demonstrate that higher certainty thresholds reduce coverage. While UQ methods have minimal impact on performance without a cut point (**Table 1**), they provide notable improvements at higher thresholds (**Table 2**). Focusing on cases where coverage exceeds 0.05, the highest AUROC is achieved by the LR model with the CP method (AUROC 0.75±0.01, AUPRC 0.74±0.01, coverage 0.36) at a cut point of 0.8. When comparing methods with equal AUPRC values, we prioritize those with superior AUROC. The XGB model with CP demonstrates the highest AUPRC of 0.82±0.02 (with AUROC of 0.67±0.04 and coverage of 6%) at the 0.9 threshold.

Across all scenarios (no cut point, 0.8, 0.9), only IR and CP methods maintain AUPRC performance without significant decreases (p-value < 0.05). Similarly, only CP methods preserve AUROC without significant degradation (p-value < 0.05). Notably, CP decreases coverage while all other uncertainty quantification methods significantly increase it (p-value < 0.05).

**Table 1**: Mean ± standard error of AUROC, AUPRC, and Coverage across 100 bootstrap results for classification models with UQ. The baseline values for UC are shown, while changes for other UQ methods are expressed as increases (+) or decreases (−) relative to UC, with (*) denoting statistical significance.

| Model | Uncertainty Method | No cut point | |
|---|---|---|---|
| | | AUROC | AUPRC |
| LR | UC | 0.68±0.0 | 0.74±0.0 |
| | PS | -0.01±0.0* | -0.0±0.0* |
| | IR | -0.01±0.0* | +0.02±0.0* |
| | VAs | -0.01±0.0* | -0.01±0.0* |
| | CP | 0.0±0.0 | 0.0±0.0 |
| SVM | UC | 0.62±0.0 | 0.7±0.0 |
| | PS | +0.01±0.0* | +0.01±0.0* |
| | IR | -0.01±0.0* | +0.07±0.0* |
| | VAs | +0.01±0.0* | +0.01±0.0* |
| | CP | 0.0±0.0 | 0.0±0.0 |
| XGB | UC | 0.65±0.0 | 0.74±0.0 |
| | PS | +0.0±0.0* | +0.0±0.0 |
| | IR | -0.01±0.0* | +0.06±0.0* |
| | VAs | +0.0±0.0 | +0.01±0.0* |
| | CP | 0.0±0.0 | 0.0±0.0 |
| RF | UC | 0.64±0.0 | 0.72±0.0 |
| | PS | +0.0±0.0* | +0.0±0.0 |
| | IR | -0.01±0.0* | +0.07±0.0* |
| | VAs | -0.0±0.0 | -0.0±0.0 |
| | CP | 0.0±0.0 | 0.0±0.0 |

**Table 2**: Mean ± standard error of AUROC, AUPRC, and Coverage across 100 bootstrap results for classification models with UQ at certainty thresholds 0.8 and 0.9. The baseline values for UC are shown, while changes for other UQ methods are expressed as increases (+) or decreases (−) relative to UC, with (*) denoting statistical significance.

| Model | Uncertainty Method | Cut point 0.8 | | | Cut point 0.9 | | |
|---|---|---|---|---|---|---|---|
| | | Coverage | AUROC | AUPRC | Coverage | AUROC | AUPRC |
| LR | UC | 0.15±0.01 | 0.71±0.02 | 0.71±0.02 | 0.04±0.0 | 0.46±0.04 | 0.65±0.03 |
| | PS | +0.4±0.01* | -0.02±0.02 | +0.03±0.02 | +0.29±0.01* | +0.23±0.04* | +0.08±0.03* |
| | IR | +0.5±0.01* | -0.02±0.02 | +0.07±0.02* | +0.44±0.01* | +0.23±0.04* | +0.15±0.03* |
| | VAs | +0.32±0.01* | -0.21±0.02* | +0.03±0.02 | +0.3±0.01* | +0.02±0.04 | +0.08±0.03* |
| | CP | +0.21±0.01* | **+0.04±0.02*** | +0.03±0.01 | +0.14±0.0* | +0.28±0.04* | +0.08±0.03* |
| SVM | UC | 0.44±0.01 | 0.61±0.01 | 0.71±0.01 | 0.21±0.01 | 0.62±0.01 | 0.71±0.01 |
| | PS | +0.27±0.01* | +0.02±0.01* | +0.01±0.0* | +0.34±0.01* | +0.02±0.01 | +0.02±0.01 |
| | IR | +0.36±0.01* | -0.0±0.01 | +0.08±0.01* | +0.51±0.01* | -0.01±0.01 | +0.09±0.01* |
| | VAs | +0.16±0.01* | -0.08±0.01* | +0.02±0.0* | +0.34±0.01* | -0.09±0.01* | +0.02±0.01 |
| | CP | -0.14±0.01* | +0.04±0.01* | +0.0±0.0 | +0.01±0.01 | +0.06±0.01* | +0.02±0.01* |
| XGB | UC | 0.49±0.01 | 0.64±0.01 | 0.77±0.01 | 0.24±0.01 | 0.6±0.02 | 0.79±0.01 |
| | PS | +0.3±0.01* | +0.02±0.01* | -0.01±0.0* | +0.41±0.01* | +0.06±0.02* | -0.03±0.01* |
| | IR | +0.36±0.01* | -0.0±0.01 | +0.05±0.01* | +0.54±0.01* | +0.04±0.02* | +0.03±0.01* |
| | VAs | +0.05±0.01* | -0.06±0.01* | +0.01±0.0* | +0.29±0.01* | -0.02±0.02 | -0.01±0.01 |
| | CP | -0.36±0.01* | +0.07±0.02* | +0.02±0.01 | -0.18±0.01* | +0.07±0.03* | **+0.03±0.02** |
| RF | UC | 0.19±0.01 | 0.59±0.02 | 0.76±0.01 | 0.05±0.0 | 0.45±0.04 | 0.81±0.02 |
| | PS | +0.44±0.01* | +0.04±0.02* | -0.02±0.01* | +0.38±0.01* | +0.17±0.04* | -0.05±0.02* |
| | IR | +0.64±0.01* | +0.03±0.02 | +0.05±0.01* | +0.72±0.01* | +0.17±0.04* | +0.01±0.02 |
| | VAs | +0.41±0.01* | -0.04±0.02* | -0.02±0.01* | +0.53±0.01* | +0.11±0.04* | -0.06±0.02* |
| | CP | -0.15±0.0* | +0.05±0.02* | +0.03±0.02 | -0.03±0.0* | -0.0±0.01 | +0.02±0.01 |

3.2 Radiomic effect on predictive model

We evaluated the impact of radiomic and dosiomic features on model performance by training three versions of the models: (1) combining radiomic and dose-based features (dosiomic + dosimetric), (2) using only dose-based features, and (3) using only dosimetric features (**Figure 3**). For discriminative performance, we calculated AUROC and AUPRC, while ACE was used to assess calibration.

As shown in **Figure 3**, incorporating a comprehensive set of features (including radiomic alongside dosiomic and dosimetric features) led to improvements in discriminative performance (AUROC and AUPRC) across most models, except in SVM that the improvement in AUPRC when compared to dosimetric and dosiomic features was not statistically significant. No model exhibited a statistically significant decrease in discriminative performance when using all features. In terms of calibration, the inclusion of all features resulted in significantly improved calibration (lower ACE), as indicated by the statistically significant negative differences in ACE for most models with exception of RF models that did not show a statistically significant improvement in calibration.

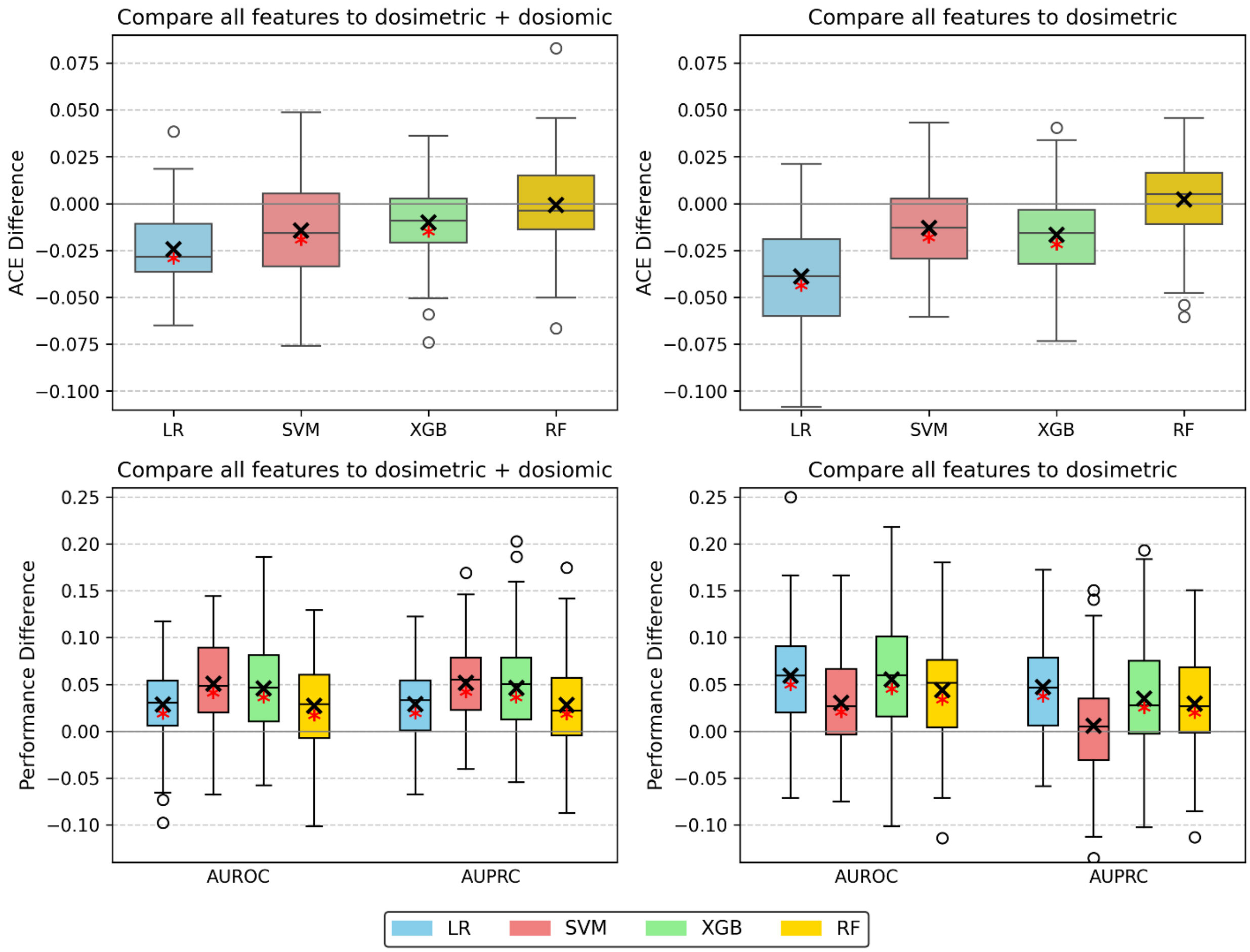

**Figure 3**: Boxplots showing differences in performance metrics between uncalibrated models trained with all features and those trained with either (1) only dosimetric features or (2) a combination of dosimetric and dosiomic features. Differences are computed as (performance with all features) minus (performance with dosimetric or dosimetric + dosiomic features). Black "X" symbols indicate the mean of each distribution, while red "*" symbols denote statistically significant mean differences. Abbreviations: LR: logistic regression; SVM: support vector machine; XGB: extreme gradient boosting; RF: random forest; ACE Adaptive Calibration Error

3.3 Effect of uncertainty quantification on calibration metrics

In this section, we evaluated the impact of UQ methods on the calibration metrics using ACE. We compare the ACE values before and after applying three calibration methods: PS, IR, and VAs. CP is not included in this analysis since it outputs p-values, which are not applicable for ACE evaluation. **Figure 4** presents the results.

The results indicate that all assessed calibration methods improve the mean ACE for the models, leading to better calibration. An exception is observed for the VAs method in the SVM and XGB models, where no significant improvement in ACE was detected. Calibration plots showing results without bootstrapping for each combination of classification model and UQ method are provided in **Supplementary Figure 3**.

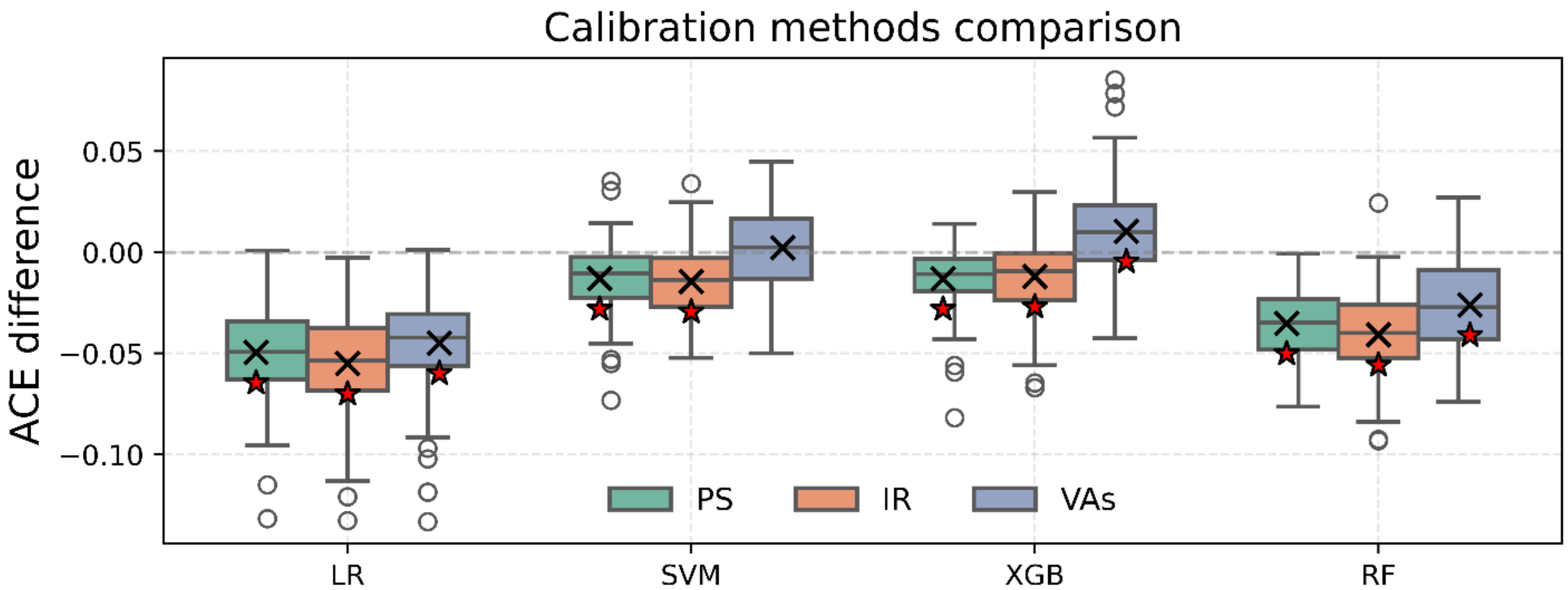

**Figure 4**: Boxplot of ACE differences between uncalibrated models and their calibrated counterparts using three calibration methods (negative values indicate improvement). Each black "X" represents the mean difference in ACE, calculated as (calibrated model) minus (uncalibrated model). A red "*" indicates statistically significant differences ($p < 0.05$). Abbreviations: LR: logistic regression; SVM: support vector machine; XGB: extreme gradient boosting; RF: random forest; UC: uncalibrated; PS: Platt scaling; IR: isotonic regression; VAs: Venn-ABERS; ACE Adaptive Calibration Error

## 4. **Discussion**

This study investigated the impact of various UQ methods on the performance and reliability of machine learning models developed for RP prediction. Given the need for reliable risk estimates in clinical decision-making and the fact that reliability is often less assessed than discriminative performance for clinical prediction models [36], we applied four distinct UQ techniques, PS, IS, VAs, and CP. These methods were evaluated across four diverse machine learning models, LR, SVM, XGB, and RF, with the resulting risk predictions and uncertainty estimate compared against an UC baseline.

Our results demonstrate that UQ can enhance certainty estimation, as discriminative performance improves when the model is confident, except for the LR model which is known for its inherent well-calibrated nature [14,37] (**Figure 2 and Table 1**). UQ methods enhance AUROC and AUPRC for the most certain predictions since they prioritize areas where the model is confident by excluding uncertain predictions that may introduce errors or noise. UQ methods operate under the premise that uncertainty correlates with error. By ranking predictions based on certainty, they exploit this correlation to identify and prioritize regions where the model is most likely to be correct. However, as coverage increases, the inclusion of uncertain predictions introduces noise and amplifies model limitations. This observation aligns with prior findings in radiomics-based locoregional recurrence prediction for head and neck cancer [20], which showed that rejecting low-certainty samples improves overall model performance.

Calibration methods such as PS, and VAs offer limited benefits in improving certainty estimates in LR (**Figure 2**). However, CP appeared to enhance uncertainty estimation for LR, as suggested by slightly improved discriminative performance (**Figure 2**) without significant performance decrease (**Table 2**), aligning with previous findings [14]. In contrast, RF, and XGB, which are more complex than LR, exhibited poor uncertainty estimation, characterized by a decrease in AUROC in regions of low certainty (low coverage levels in **Figure 2**). This unreliability in initial uncertainty from models can be mitigated through UQ techniques (**Figure 2, Table 2**). Similarly, SVM in this study generally maintained performance in low-certainty regions but also showed potential for performance improvement through UQ. This can be attributed to the fact that increased model complexity often introduces overfitting and unreliable probability estimates, resulting in worsened calibration performance [38,39]. For instance, SVM and XGB can exacerbate calibration issues by assigning overly confident probabilities to outliers or misclassified points, further skewing their output reliability. Meanwhile, calibration issues in RF arise due to its ensemble structure (bagging ensemble), which averages predictions from decision trees and produces unreliable probabilities, particularly near class boundaries [37]. This results in poor estimates as RF struggles to model the smooth transitions in probability distributions necessary for well-calibrated outputs. Additionally, prior research similarly reports poorer calibration in SVM, RF [37] and XGB [40] compared to LR.

Regarding the impact of features, our results generally suggest that incorporating radiomic and/or dosiomic features alongside dosimetric features significantly improved both discriminative performance and calibration (**Figure 3**). While the high-dimensional nature of radiomic features can sometimes raise concerns about generalizability and overfitting [41], our findings suggest reduced overfitting when these features are included. Nevertheless, complex models (SVM, RF and XGB) appeared to exhibit a smaller reduction in ACE compared to the LR model when these features were included (**Figure 3**). This outcome may indicate that the methodology employed in this study helped mitigate overfitting effects on calibration.

Compared to previous studies on RP prediction, our results are in line with most radiomic-based approaches. A recent review summarizing 16 such studies reported an overall AUROC of approximately 0.76 [3]. One study, which integrated clinical, dosimetric, dosiomic, and radiomic features using LR with L1 norm regularization in lung cancer patients treated with stereotactic body radiation therapy, achieved an AUROC of 0.77 [5]. This is close to the AUROC of $0.75 \pm 0.01$ obtained in our study using LR with CP. Notably, one study employing extremely randomized trees (similar to RF model) with a similar combination of features (dosimetric, dosiomic, and radiomic) reported a higher AUROC of 0.92 [42]. It is important to note, however, that direct comparisons between studies remain challenging due to variations in study design, treatment modalities, and patient populations. These differences may influence model performance.

In summary, our findings indicate that for clinical use, UQ techniques should be applied especially to complex ML models, such as SVM, RF, and XGB, to enhance the reliability of their predictions. LR, a less complex model, often achieves comparable or superior performance [36,43] with greater reliability in probability estimates. CP demonstrated effectiveness in the overall improvement of reliability on certain predictions. Furthermore, radiomic and dosiomic features alongside dosimetric features significantly improved both the models' discriminative power and calibration.

## Declaration of interests

The authors declare that they have no known competing financial interests or personal relationships that could have appeared to influence the work reported in this paper.

## Declaration of Generative AI and AI-assisted technologies in the writing process

During the preparation of this work the author(s) used ChatGPT in order to improve readability and language. After using this tool/service, the author(s) reviewed and edited the content as needed and take(s) full responsibility for the content of the publication

## Supplementary material

### A. Dataset

Initially, 336 esophageal cancer patients were identified, but 235 were excluded according to lacked treatment planning data, had a history of malignancy or prior radiation therapy, had underlying interstitial lung disease, had a follow-up period of less than one year, developed lung metastases within a year, or were treated with brachytherapy. Ultimately, 101 patients were eligible and included in the final analysis (**Supplementary Figure 1**).

In this study, the National Cancer Institute Common Terminology Criteria for Adverse Events version 5.0 (CTCAE v5.0) was used as the basis for grading radiation pneumonitis, with the following classifications:

- Grade 0: No symptoms and no radiographic features
- Grade 1: Mild symptoms not requiring steroids, or radiographic features only
- Grade 2: Symptoms interfering with daily activities, or requiring steroid treatment
- Grade 3: Symptoms requiring both steroids and oxygen therapy
- Grade 4: Symptoms requiring intubation

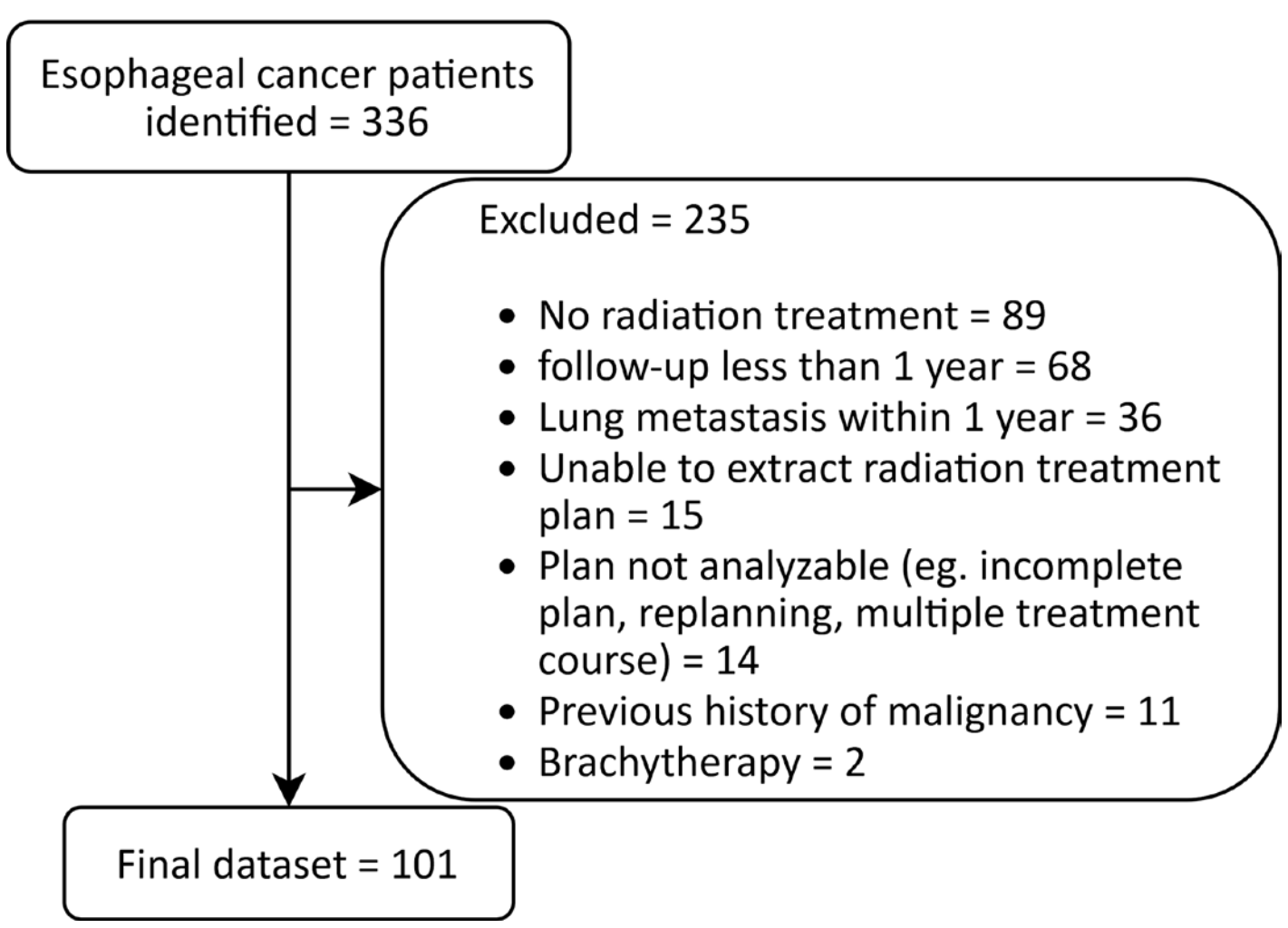

**Supplementary Figure 1:** Consort Diagram

**Supplementary Table 1:** Patient characteristics (N=101)

| **Parameters** | **Median (Range)/N (%)** |
|---|---|
| **Age** | 61 (26-93) |
| **Gender** | |
| Male | 89 (88%) |
| Female | 12 (12%) |
| **Smoking status** | |
| Never | 29 (29%) |
| Active | 25 (25%) |
| Quit smoking < 10 years | 33 (33%) |
| Quit smoking > 10 years | 14 (13%) |
| **ECOG performance status** | |
| 0 | 32 (32%) |
| 1 | 60 (60%) |
| 2 | 9 (8%) |
| **Stage** | |
| 1 | 4 (4%) |
| 2 | 3 (3%) |
| 3 | 71 (70%) |
| 4 | 23 (23%) |
| **Chemotherapy regimen** | |
| Cisplatin + 5-FU | 23 (23%) |
| Carboplatin + 5-FU | 8 (8%) |
| Carboplatin + paclitaxel | 60 (59%) |
| Carboplatin alone | 2 (2%) |
| Paclitaxel alone | 2 (2%) |
| No chemotherapy | 6 (6%) |

**Supplementary Table 2:** Treatment characteristics (N=101)

| **Parameters** | **Median (Range)/N (%)** |
|---|---|
| **Surgery** | |
| Yes | 26 (26%) |
| No | 75 (74%) |
| Prescription dose | 50.4 (30.0-60.0) |
| Prescription dose per fraction | 1.8 (1.8-3.0) |
| **RT technique** | |
| 3D conformal | 78 (77%) |
| IMRT/VMAT | 9 (9%) |
| Combine | 14 (14%) |
| **RT setting** | |
| Preoperative | 47 (47%) |
| Postoperative | 1 (1%) |
| Definitive | 49 (48%) |
| Palliative | 4 (4%) |
| **RP grade** | |
| 0 | 38 (38%) |
| 1 | 58 (57%) |
| 2 | 5 (5%) |
| 3 | 0 (0%) |
| **Dosimetric Parameter (Lung)** | **Mean (Range)** |
| MLD (Gy) | 10.3 Gy (1.1-16.3 Gy) |
| V5 (%) | 48.9% (3.4-74.0%) |
| V10 (%) | 32.0% (2.8-53.0%) |
| V20 (%) | 16.2 (2.1-31.5%) |
| V30 (%) | 11.1 (1.6-26.0%) |
| V40 (%) | 6.2% (0.0-18.9%) |

### B. Preprocessing and features

**Preprocessing**

Prior to radiotherapy, pretreatment CT images were acquired using the Optima 580 CT simulator (GE Healthcare, Milwaukee, WI, USA) prior to radiotherapy treatment with free-breathing technique. The acquisition parameters included a tube voltage of 120 kVp, and the reconstruction algorithm was filtered back projection. All radiation dose distributions were then calculated using the Analytical Anisotropic Algorithm within the Eclipse Treatment Planning System. The total dose distribution was converted to an equivalent dose of 2 Gy (EQD2) using the following formula: $EQD2_k = \sum_i^N \frac{d_{I,k}+d_{i,k}^2/(\alpha/\beta)}{1+2/(\alpha/\beta)}$, where $d_{i,k}$ is dose at fraction $i$ and voxel $k$. The value of $\alpha/\beta$ was set to 3. From now on the dose distribution will refer to dose distribution in EQD2. The dose distributions and pretreatment CT images were then resampled to have voxel size of 1.5 × 1.5 × 1.5 mm3 using b-spline algorithm. The regions of interests (ROIs) were resampled to match the pretreatment CT images using the nearest neighbor algorithm.

**Anti-aliasing filter**

Before resampling the CT image, we applied a Gaussian filter as an anti-aliasing measure, following a similar approach to previous studies [1,2]. The purpose of this filter was to remove spatial frequencies higher than the Nyquist frequency ($w_N$) which is calculated by $w_N = \frac{d_1}{2d_2}$, for $d_1$ is original in-plane voxel spacing (around 1 mm) and $d_2$ is the desired voxel spacing (1.5 mm). The standard deviation (**σ**) of the Gaussian kernel was calculated using the formula:

$$\sigma = \frac{1}{w_N}\sqrt{-2\ ln(\beta)}$$

where $\beta$ is the filter's magnitude response at the Nyquist frequency. We set $\beta$ = 0.98 [2] in this study. The dose map was resampled without applying the Gaussian filter.

**Features extraction**

In this study, we extracted three categories of features: dosimetric, dosiomic (derived from the dose distribution), and radiomic (derived from the pretreatment CT images). Dosimetric features included mean lung dose and relative lung volume receiving more than a specific dose threshold x (Vx), for x in [5, 10, …, 70]. Dosiomic and radiomic features were extracted using the PyRadiomics v3.0.1.

The dosiomic feature set included 18 first-order statistics and 61 texture features: 24 from the gray-level co-occurrence matrix (GLCM), 16 from the gray-level run length matrix (GLRLM), 16 from the gray-level size zone matrix (GLSZM), and 5 from the neighborhood gray tone difference matrix (NGTDM). These were computed from the EQD2 dose distribution, discretized using a fixed bin width of 1 Gy over the 0–70 Gy range. Radiomic features were extracted from pretreatment CT images within three dose-based lung ROIs, defined by thresholds of 10 Gy, 15 Gy, and 20 Gy, similar to previous study. Prior to extraction, Hounsfield units (HUs) were clipped to the range [-1000, 100] and then linearly rescaled to [0, 1100]. The same set of first-order and texture features used for dosiomics was extracted for each ROI, resulting in three times the number of radiomic features relative to dosiomic ones. Radiomic feature extraction used a fixed bin width of 50 HU over the 0–1100 range. Unless otherwise specified, all other feature extraction parameters followed the default settings of PyRadiomics v3.0.1.

### C. Hyperparameters

This supplementary material details the hyperparameter tuning process and model selection strategy employed in our study. We utilized a 3-fold cross-validation (CV) approach combined with a grid search to optimize hyperparameters for each model. The specific hyperparameters explored for each model, along with their corresponding values, are listed in Supplementary Table 1. Hyperparameters not listed were left at their default values as implemented in the respective libraries.

**Supplementary table 3**: Hyperparameters used in this study. Hyperparameters not listed were left at their default values.

| Model | Library | Hyperparameters | Values |
|---|---|---|---|
| Logistic Regression (LR) | scikit-learn | penalty | l2, l1 |
| | | C | 1, 0.1, 10, 100, 0.01 |
| | | solver | liblinear |
| | | class_weight | balanced |
| | | max_iter | 2000 |
| Random Forest (RF) | scikit-learn | n_estimators | 5, 20, 50 |
| | | min_samples_leaf | 2, 4 |
| | | min_samples_split | 2, 4 |
| | | class_weight | balanced |
| Support Vector Machine (SVM) | scikit-learn | C | 1, 0.1, 10, 0.01 |
| | | kernel | rbf |
| | | gamma | scale |
| | | class_weight | balanced |
| | | probability | TRUE |
| XGBoost (XGB) | xgboost | n_estimators | 5, 20, 50 |
| | | learning_rate | 0.1, 0.01 |
| | | subsample | 0.5, 1 |
| | | scale_pos_weight | Calculated based on class distribution for each fold |

**D. Training process**

Here's an example to illustrate the training process described in Figure 1 of the manuscript. Let's imagine we have a very small dataset with just three samples: Sample X, Sample Y, and Sample Z. We'll consider two pipelines (combinations of a classification model and an uncertainty quantification method) and two bootstrap iterations for simplicity.

Assumptions

1. Our Pipelines:
   - Pipeline 1: Logistic Regression (LR) with Uncalibrated uncertainty (UC)
   - Pipeline 2: Support Vector Machine (SVM) with Prediction Scores (PS)
2. Our Bootstrap Iterations: 2

Leave-One-Out cross validation (LOO-CV) Process:

The LOO-CV will run three times, with each sample serving as the test sample once.

Iteration 1: Sample X is the test sample, Training set is {Y, Z}

- Pipeline 1 (LR + UC):
  - Bootstrap 1: Train LR on a bootstrapped version of {Y, Z}. Predict for X: prediction_X_pipeline1_bootstrap1
  - Bootstrap 2: Train LR on another bootstrapped version of {Y, Z}. Predict for X: prediction_X_pipeline1_bootstrap2
- Pipeline 2 (SVM + PS):
  - Bootstrap 1: Train SVM on a bootstrapped version of {Y, Z}. Predict for X: prediction_X_pipeline2_bootstrap1
  - Bootstrap 2: Train SVM on another bootstrapped version of {Y, Z}. Predict for X: prediction_X_pipeline2_bootstrap2

Iteration 2: Sample Y is the test sample, Training set is {X, Z}

- Pipeline 1 (LR + UC):
  - Bootstrap 1: Train LR on a bootstrapped version of {X, Z}. Predict for Y: prediction_Y_pipeline1_bootstrap1
  - Bootstrap 2: Train LR on another bootstrapped version of {X, Z}. Predict for Y: prediction_Y_pipeline1_bootstrap2
- Pipeline 2 (SVM + PS):
  - Bootstrap 1: Train SVM on a bootstrapped version of {X, Z}. Predict for Y: prediction_Y_pipeline2_bootstrap1
  - Bootstrap 2: Train SVM on another bootstrapped version of {X, Z}. Predict for Y: prediction_Y_pipeline2_bootstrap2

Iteration 3: Sample Z is the test sample, Training set is {X, Y}

- Pipeline 1 (LR + UC):
    - Bootstrap 1: Train LR on a bootstrapped version of {X, Y}. Predict for Z: prediction_Z_pipeline1_bootstrap1
    - Bootstrap 2: Train LR on another bootstrapped version of {X, Y}. Predict for Z: prediction_Z_pipeline1_bootstrap2
- Pipeline 2 (SVM + PS):
    - Bootstrap 1: Train SVM on a bootstrapped version of {X, Y}. Predict for Z: prediction_Z_pipeline2_bootstrap1
    - Bootstrap 2: Train SVM on another bootstrapped version of {X, Y}. Predict for Z: prediction_Z_pipeline2_bootstrap2

Aggregation:

Now, for each combination of pipeline and bootstrap iteration, we aggregate the predictions made for each test sample in the LOO-CV process.

- Pipeline 1 (LR + UC) - Bootstrap 1: [prediction_X_pipeline1_bootstrap1, prediction_Y_pipeline1_bootstrap1, prediction_Z_pipeline1_bootstrap1]
- Pipeline 1 (LR + UC) - Bootstrap 2: [prediction_X_pipeline1_bootstrap2, prediction_Y_pipeline1_bootstrap2, prediction_Z_pipeline1_bootstrap2]
- Pipeline 2 (SVM + PS) - Bootstrap 1: [prediction_X_pipeline2_bootstrap1, prediction_Y_pipeline2_bootstrap1, prediction_Z_pipeline2_bootstrap1]
- Pipeline 2 (SVM + PS) - Bootstrap 2: [prediction_X_pipeline2_bootstrap2, prediction_Y_pipeline2_bootstrap2, prediction_Z_pipeline2_bootstrap2]

Each of these lists will be compared to the corresponding true labels for our samples, which we can represent as y_true = [true_label_X, true_label_Y, true_label_Z]. For instance, for the first list, we would compare prediction_X_pipeline1_bootstrap1 with true_label_X, prediction_Y_pipeline1_bootstrap1 with true_label_Y, and prediction_Z_pipeline1_bootstrap1 with true_label_Z. This comparison allows us to calculate various evaluation metrics for that specific combination of pipeline and bootstrap iteration. As the example shows, for each pipeline we considered (Pipeline 1: LR + UC, and Pipeline 2: SVM + PS), we obtained two bootstrap results. This is because we ran the bootstrapping process twice for each pipeline.

The total number of unique "pipelines" (defined as a specific combination of a classification model and an uncertainty quantification method) of the actual study is the product of the number of models and the number of uncertainty methods = 4 × 5 = 20 pipelines. The study's workflow was designed to train and evaluate each of these 20 different pipelines across 100 bootstrap iterations. In the actual study described in **Figure 1** in the main manuscript, the training set is bootstrapped 100 times, so for each pipeline, you would have 100 bootstrap results, each with its own set of aggregated predictions and corresponding evaluation metrics.

### E. Uncertainty quantification methods

Given a dataset $\{(x_1, y_1), \ldots, (x_n, y_n)\}$, where $x_i$ represents the feature vector for the $i^{th}$ instance and $y_i$ represents the corresponding label, we focus on a binary classification problem. The uncertainty score, denoted as $s(x)$, is defined as:

$$s(x) = \begin{cases} 1 - p(x), & \hat{y} \text{ is positive} \\ p(x), & \hat{y} \text{ is negative} \end{cases}$$

Here, $p(x)$ represents the predicted probability or p-value of the positive class, potentially with or without the application of UQ methods (raw output from classifier, denoted as $f(x)$). This formulation assumes that a prediction $\hat{y}$ is made based on a threshold of $p(x) \geq 0.5$ ($f(x) \geq 0.5$ if CP since CP output p-value). Consequently, the model exhibits the highest certainty when the predicted probability/p-value is either 0 or 1. Below are detailed descriptions of each uncertainty quantification method:

- UC: The UC method simply refers to using the output probabilities from the classifier without applying any UQ techniques. Thus, $p(x)$, is equivalent to $f(x)$.
- PS [3]: PS is a parametric calibration method that fits LR to the classifier's output scores, converting them into calibrated probabilities. The process starts by applying a classifier without UQ to obtain a dataset of output-label pairs $\{(f(x_1), y_1), \dots, (f(x_m), y_m)\}$, where $f(x_i)$ represents the raw output score from the classifier, and $y_i \in \{0,1\}$ is the corresponding true label in a binary classification. These output-label pairs are then used to fit a LR defined by:

$$p\big(f(x)\big) = \frac{1}{1 + exp\big(a + bf(x)\big)}$$

  where $a$ and $b$ are learned parameters by minimizing the negative log-likelihood of the observed data. This method ensures that the predicted probabilities lie within the range [0, 1] and are more reliable for interpreting uncertainty.

- IR: IR is a non-parametric calibration technique that fits the data to a piecewise constant function, subject to the constraint that the function is non-decreasing. The optimization process involves minimizing the root mean square error between predicted probabilities $p(x_i)$ and actual outcomes $y_i$.
- VAs [4]: VAs is a calibration technique based on Venn prediction, which provides multiple probability estimates instead of a single probability by applying two IR models to the raw output of a classifier. It consists of two steps: first, IR is fitted to the probability of being a positive class using the training set $\{(f(x_1), y_1), \dots, (f(x_m), y_m)\}$ along with a test sample $(f(x_i), 1)$. Second, IR is fitted to the probability of being a negative class using the training set $\{(1 - f(x_1), y_1), \dots, (1 - f(x_m), y_m)\}$ along with a test sample $(f(x_i), 0)$. The first IR model computes $p_1(x_i)$, the probability of $x_i$ belonging to class 1 given $f(x_i)$ while the second IR model computes $p_0(x_i)$, the probability of $x_i$ belonging to class 0 given $f(x_i)$. In practice, multiple probabilities will be merge to $p(x)$ by:

$$p(x) = \frac{p_1}{1 - p_0 + p_1}.$$

- CP [5–7]: CP provides a framework for generating prediction sets for any model. It works by evaluating how well a new prediction aligns with the distribution of previously observed data, using insights learned from the training set. To compute p-value within CP framework, we first calculate the nonconformity score, denoted as $\alpha_i$, for a given data $x_i$. Nonconformity score $\alpha_i$ is defined as $-log\, f(x_i)$ [14,30]. This score measures how "non-conforming" the test data is compared to the training data distribution. Once the nonconformity scores are obtained, they are used to compute the p-value, $p(x)$, which reflects how the test point is deviated relative to the training set. Specifically, the p-value is calculated as:

$$p(x_{m+1}) = \frac{\#\{i = 1, \cdots, m+1 \mid \alpha_i \geq \alpha_{m+1}\}}{m+1} \text{ [8,9].}$$

Where $\{\alpha_1, \cdots, \alpha_m\}$ represents the set of nonconformity score from the training dataset and $\alpha_{m+1}$ is the nonconformity score of the test data, $\#\{\}$ indicates the number of elements in the set.

### F. Adaptive calibration error (ACE)

Given number of classes ($K$), number of samples ($N$) and number of ranges ($R$), ACE is defined as:

$$ACE = \frac{1}{KR}\sum_{k=1}^{K}\sum_{r=1}^{R} | \, acc(r,k) - conf(r,k) \, |$$

where $acc(r,k)$ and $conf(r,k)$ are the accuracy and confidence of adaptive calibration range $r$ for class label $k$, respectively. The calibration range $r$ is determined by the $\lfloor N/R \rfloor^{\text{th}}$ index of the sorted probability output.

### G. Results without bootstrap

In addition to the workflow involving bootstrap resampling, we also conducted experiments where the classification models were trained on the entire training set within each Leave-One-Out Cross-Validation (LOO-CV) split, without any bootstrapping. This approach resulted in a single prediction for each test sample in each LOO-CV iteration for every combination of model and uncertainty quantification method. As there are four classification models (LR, SVM, XGB, RF) and five uncertainty quantification methods (UC, PS, IR, VAs, CP), this resulted in a total of 20 sets of results.

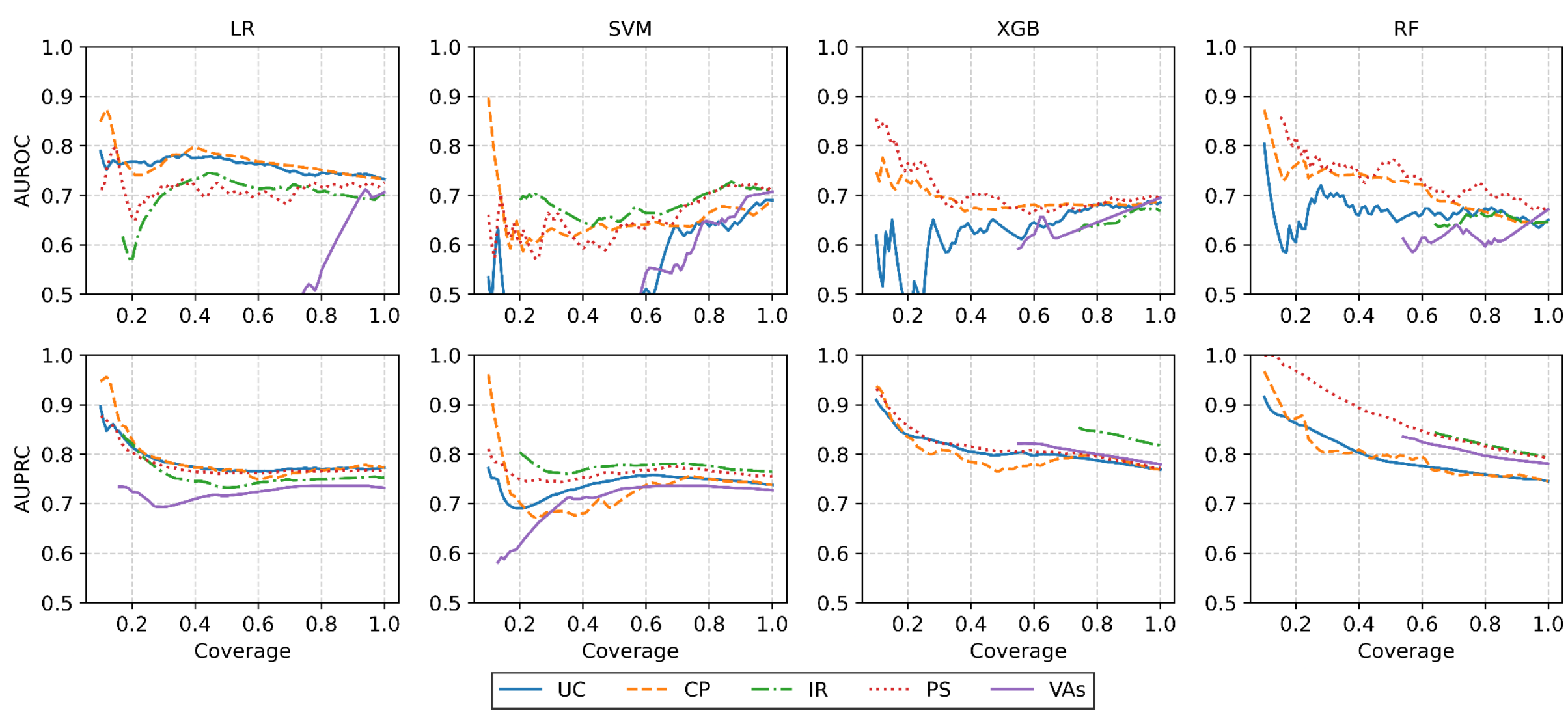

**Supplementary figure 2**: AUROC and AUPRC of each ML model with different UQ methods across varying coverage levels.

**Supplementary Table 4**: This table presents AUROC and AUPRC for each classification model combined with UQ methods across different certainty thresholds (no threshold, 0.8 and 0.9). The baseline values for UC are shown, while changes for other UQ methods are expressed as increases (+) or decreases

(−) relative to UC. (*N/A indicates that no data were selected for evaluation at that threshold and improvements for UQ methods are calculated using 0 as the baseline)

| Model | Uncertainty Method | No Cut point | | Cut point 0.8 | | | Cut point 0.9 | | |
|---|---|---|---|---|---|---|---|---|---|
| | | AUROC | AUPRC | Coverage | AUROC | AUPRC | Coverage | AUROC | AUPRC |
| LR | UC | 0.73 | 0.77 | 0.08 | 0.8 | 0.92 | 0.03 | 0.0 | 0.17 |
| | PS | -0.01 | -0.01 | +0.34 | -0.09 | -0.16 | +0.17 | +0.66 | +0.64 |
| | IR | -0.03 | -0.02 | +0.47 | -0.08 | -0.19 | +0.18 | +0.59 | +0.65 |
| | VAs | -0.03 | -0.04 | +0.33 | -0.53 | -0.21 | +0.14 | +0.5 | +0.57 |
| | CP | 0.0 | 0.0 | +0.29 | -0.01 | -0.14 | +0.21 | +0.74 | +0.64 |
| SVM | UC | 0.69 | 0.74 | 0.09 | 0.65 | 0.81 | 0.0 | N/A* | N/A* |
| | PS | +0.02 | +0.02 | +0.41 | -0.04 | -0.05 | +0.21 | +0.63* | +0.75* |
| | IR | +0.02 | +0.03 | +0.48 | +0.03 | -0.04 | +0.23 | +0.7* | +0.79* |
| | VAs | +0.02 | -0.01 | +0.42 | -0.27 | -0.09 | +0.15 | +0.45* | +0.59* |
| | CP | 0.0 | 0.0 | +0.16 | -0.05 | -0.14 | +0.17 | +0.59* | +0.72* |
| XGB | UC | 0.69 | 0.77 | 0.43 | 0.64 | 0.8 | 0.23 | 0.51 | 0.83 |
| | PS | +0.01 | +0.0 | +0.31 | +0.03 | +0.0 | +0.38 | +0.16 | -0.03 |
| | IR | -0.02 | +0.05 | +0.44 | +0.01 | +0.04 | +0.54 | +0.13 | +0.01 |
| | VAs | +0.01 | +0.01 | +0.19 | +0.01 | +0.02 | +0.32 | +0.08 | -0.01 |
| | CP | 0.0 | 0.0 | -0.2 | +0.09 | +0.01 | -0.12 | +0.21 | +0.1 |
| RF | UC | 0.65 | 0.75 | 0.21 | 0.65 | 0.86 | 0.04 | 0.0 | 1.0 |
| | PS | +0.03 | +0.05 | +0.42 | +0.06 | -0.02 | +0.47 | +0.77 | -0.13 |
| | IR | -0.0 | +0.05 | +0.56 | +0.01 | -0.04 | +0.63 | +0.64 | -0.16 |
| | VAs | +0.02 | +0.04 | +0.41 | -0.04 | -0.04 | +0.52 | +0.59 | -0.17 |
| | CP | 0.0 | 0.0 | 0.0 | +0.12 | +0.01 | +0.02 | 0.0 | 0.0 |

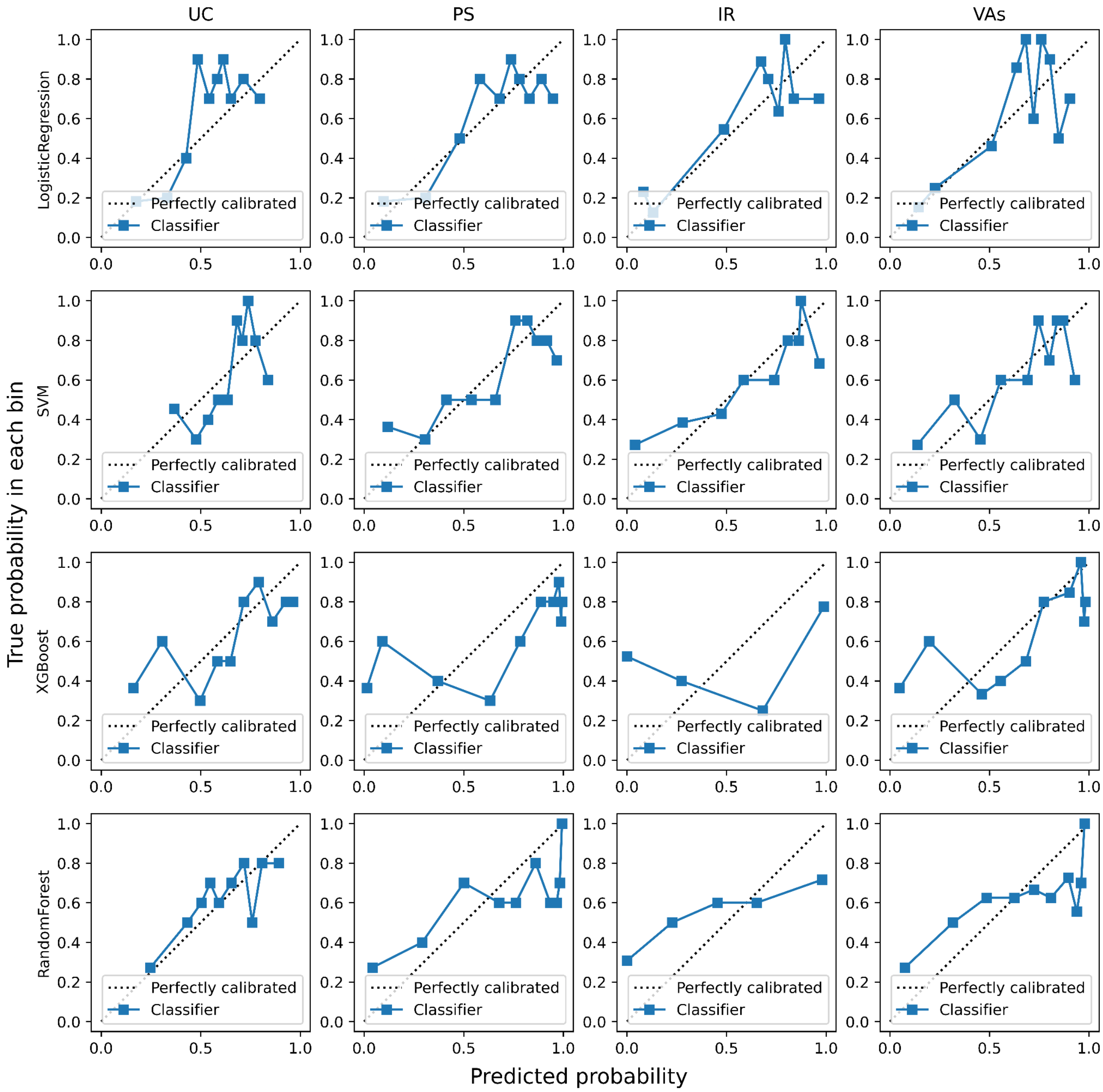

**Supplementary Figure 3**: Calibration plots. Each panel shows the calibration curve for a specific classifier trained on all feature types (dosimetric, dosiomic and radiomic). The x-axis represents the predicted probability output by the classifier, and the y-axis represents the true probability of the positive class within bins of predicted probabilities. The solid blue line with markers shows the calibration of the classifier, while the dotted diagonal line represents perfect calibration. Rows correspond to different classifiers, and columns correspond to different UQ methods.